\pdfoutput=1

\documentclass[11pt]{article}

\usepackage[]{acl}
\usepackage{wasysym}
\usepackage{soul}
\usepackage{color}
\newcommand{\hlc}[2][yellow]{{\sethlcolor{#1}\hl{#2}}}

\usepackage{times}
\usepackage{latexsym}

\usepackage[T1]{fontenc}

\usepackage[utf8]{inputenc}

\usepackage{microtype}

\usepackage{graphicx}
\graphicspath{ {./img/} }

\title{EmoMent: An Emotion Annotated Mental Health Corpus from two South Asian Countries}

\author{Thushari Atapattu$^1$,  Mahen Herath$^2$, Charith Elvitigala$^3$, Piyanjali de Zoysa$^4$,\\ Kasun Gunawardane$^3$, Menasha Thilakaratne$^1$,\\ Kasun de Zoysa$^3$ \and Katrina Falkner$^1$ \\ $^1$School of Computer Science, The University of Adelaide, Adelaide, Australia \\ $^2$Department of Computer Science \& Engineering, University of Moratuwa, Katubedda, Sri Lanka \\ $^3$University of Colombo School of Computing, Colombo, Sri Lanka \\ $^4$Department of Psychology, University of Colombo, Sri Lanka \\ email: thushari.atapattu@adelaide.edu.au}

\begin{document}
\maketitle
\begin{abstract}
People often utilise online media (e.g., Facebook, Reddit) as a platform to express their psychological distress and seek support. State-of-the-art NLP techniques demonstrate strong potential to automatically detect mental health issues from text. Research suggests that mental health issues are reflected in \textit{emotions} (e.g., sadness) indicated in a person's choice of language. Therefore, we developed a novel emotion-annotated mental health corpus (\textit{EmoMent}), consisting of 2802 Facebook posts (14845 sentences) extracted from two South Asian countries - Sri Lanka and India. Three clinical psychology postgraduates were involved in annotating these posts into eight categories, including `mental illness' (e.g., \textit{depression}) and emotions (e.g., `\textit{sadness}', `\textit{anger}'). EmoMent corpus achieved `very good' inter-annotator agreement of 98.3\% (i.e. \% with two or more agreement) and Fleiss' Kappa of 0.82.
Our RoBERTa based models achieved an F1 score of 0.76 and a macro-averaged F1 score of 0.77 for the \textit{first task} (i.e. predicting a mental health condition from a post) and the \textit{second task} (i.e. extent of association of relevant posts with the categories defined in our taxonomy), respectively.
\end{abstract}

\section{Introduction}

Mental health issues remain a leading cause for poor well-being and suicide. The World Health Organisation (WHO) indicates that 400 million people are affected by mental disorders such as depression, resulting in a cost of US\$ 1 trillion per year from the global economy allocated for depression and anxiety disorders alone \cite{who1,james2018global}. Recent research using AI and NLP demonstrates strong potential to automatically detect mental health issues from digital footprints such that professionals could provide timely interventions and mental health resources to vulnerable persons. 
These data contain useful information to understand patients' distressed state of mind outside a traditional clinical environment. 

Research suggests that mental health issues are reflected in the `emotions' (e.g., sadness, anger) indicated in one's expression of language. Despite the popularity of research studies in detecting mental disorders using online data such as \textit{Twitter} \cite{coppersmith2014quantifying,coppersmith2015clpsych,cohan2018smhd} and emotion modeling \cite{strapparava2007semeval,mohammad2018semeval,demszky2020goemotions,oberlander2018analysis}, the automated identification of the \textit{association between emotions and mental disorders} have largely being ignored, apart from a recent study (CEASE corpus \cite{ghosh2020cease}) that focused on the role of emotions on suicidal ideation. 

Motivated by this, we introduce a novel, emotion-annotated mental health (\textit{EmoMent}) corpus\footnote{dataset and the code is available on request for research purposes.} using \textit{Facebook} posts
extracted from two South Asian countries - Sri Lanka and India. 
In South Asia, due to the lack of awareness of symptoms of mental illnesses and its associated stigma, people often do not seek professional help, resulting in many instances of mental disorders being left undiagnosed \cite{arora2016attitudes}.
However, since recently, these countries have demonstrated a tendency to use social media, particularly Facebook, to seek mental health help using private and public groups (e.g., \textit{Psychology group} in Sri Lanka, \textit{Indian Psychology Association}).  

Depression and anxiety disorders are amongst the most common mental disorders worldwide \cite{james2018global,blackdog}. Therefore, our dataset includes de-identifiable Facebook posts from individuals who have indicated a diagnosis of depression
or anxiety, the disorder-related issues they express including associated emotions, and their help-seeking behaviours from professionals and/or community. 
EmoMent consists of 2802 posts (14845 sentences) extracted from public Facebook groups dedicated to discuss mental health concerns in Sri Lanka and India. Three clinical psychology postgraduates were involved in the data annotation process. Their task was to read the entire post and assign one or more labels from a given set of eight categories (e.g., \textit{mental illness, sadness, psychosomatic, irrelevant}) (Table \ref{tab:taxonomy_example}). We have achieved `very good' inter-annotator agreement of 98.3\% (i.e. \% with two or more rater-agreement) and Fleiss' Kappa of 0.82, while 0.90 and 0.74 of Kappa values were achieved on Sri Lankan and Indian datasets respectively, enabling a promising human agreement for computational modelling.

We fine-tuned BERT \cite{devlin2018bert} and RoBERTa \cite{liu2019roberta} based deep learning models on the EmoMent corpus to predict the relevance of a post to a mental health condition (\textit{first task)}, and to associate relevant posts with the categories defined in our taxonomy (\textit{second task}) (see section 3.3). Our RoBERTa-based models achieved a F1 score of 0.76 for the \textit{first task}, and a macro-averaged F1 score of 0.77 for the \textit{second task}.


The novel contributions of our paper includes; 1) the development of the first emotion-annotated mental health corpus in English language 2) the development of the first \textit{taxonomy} to annotate mental health conditions and emotions from Facebook data, and 3) the development and evaluation of deep learning models (RoBERTa) to predict the presence of mental conditions, emotions, and psychosomatic issues with `good' performance. Additionally, our research contributed to the integration of knowledge from two domains - \textit{mental health} and \textit{emotion modelling} through various quantitative and qualitative analyses, in particular, low-resource languages such as Sinhala.   

Currently, the diagnosis of a mental disorder is primarily based on the knowledge and experience of a professional, who arrives at a diagnosis subsequent to talking to a patient and/or care-givers. In this method, patients have to reflect on events that occurred in the past 
to help professionals diagnose their condition. Real-time experiences of patients, which is an important element for diagnosis and treatment plan, is not usually considered. The majority of online self-reflective posts on the other hand generate real-time, reliable data to uncover distressed states of mind at the time of occurring. Therefore, a corpus like EmoMent, developed from user-generated data allows practitioners to understand the mental states of patients beyond a traditional clinical interview. These automated identification of mental disorders or mental conditions,from user-generated content provides a useful tool for improving diagnosis and personalised treatment plans. 
\section{Related Work}

\label{sec:litrature}
People use language as a direct tool to express their feelings and emotions, providing a wealth of information to determine their emotional status and mental health conditions \cite{berry2017whywetweetmh}. Motivated by this, many datasets have been developed to support research in the two fields of: \textit{Emotion Modelling} and \textit{Mental Health Modelling}, using social media as one of the primary data sources. The existing datasets on \textit{emotion modelling} are mostly based on two emotion taxonomies:  Ekman's basic emotions (\textit{fear}, \textit{anger}, \textit{joy}, \textit{sadness}, \textit{disgust}, and \textit{surprise}) \cite{ekman1992argument}, and Plutchik's Wheel of Emotions (\textit{anger}, \textit{anticipation}, \textit{disgust}, \textit{fear}, \textit{joy}, \textit{sadness}, \textit{surprise} and \textit{trust}) \cite{plutchik1980general}. Examples of emotion modelling datasets include \cite{strapparava2007semeval,mohammad2018semeval, demszky2020goemotions, oberlander2018analysis, li2020we, appidi2020creation}. The existing \textit{mental health modelling} datasets are based on the problem domains such as suicidal attempts, self-injury, loneliness, depression, anxiety and Post Traumatic Stress Disorder. The focus of most existing datasets are limited to one or two  problem domains, hindering the diagnosis capabilities of the AI models that they are based on. \cite{pirina2018identifying,tadesse2019detection,zirikly2019clpsych}.  

Despite the availability of numerous emotion modelling and mental health modelling datasets, there are certain limitations in almost all of these datasets. The datasets from these two research fields are independent of one another (i.e., non-interactive). The complementary integration of emotion and mental health modelling provide enhanced insights on a person's emotional and mental well-being, which is useful in assisting professionals to diagnose and personalise treatment plans. Additionally, almost all of the currently available datasets are based on resource-rich languages such as English \cite{appidi2020creation}, limiting the understanding of cultural aspects of language use in emotion and mental health modelling.

Despite the importance of jointly-modelling emotions and mental health, the availability of emotion-annotated mental health datasets are vastly limited. For example, the emotion-annotated mental health dataset of Ghosh et. al. \cite{ghosh2020cease}, CEASE, is specific to suicide notes. Motivated by this, we present a novel emotion-annotated mental health dataset based on Facebook data to facilitate joint-modeling of emotions and mental health conditions. 

To propose baseline models from the constructed datasets, most previous studies in emotion, mental health, and emotion-annotated mental health domains have utilised recent advancements of deep learning techniques such as BERT, LSTMs and RNNs \cite{li2020we,appidi2020creation}. For example, the CEASE dataset proposes an ensemble model using LSTM, CNN, and GRU \cite{ghosh2020cease}. To adhere with this, we also leveraged the recent advancements in deep learning techniques using BERT and RoBERTa based models.


\section{EmoMent Corpus}
This section describes the development of the EmoMent corpus - data collection, data cleaning, taxonomy development, and data annotation.

\subsection{Data Collection}
We used the CrowdTangle tool\footnote{https://www.crowdtangle.com} to collect Facebook posts that express mental health-related issues. CrowdTangle is a content discovery and social monitoring platform which provides an interface to access \textit{public} Facebook pages and group posts. Their search interface contains filters such as 
`Post type - photos, statuses', `language', and `time frame'. 
Our search filter parameters were `account type' as groups and `post type' as statuses. Our `language' parameters were \textit{Sinhala} and \textit{English} while restricting `geographical locations' to Sri Lanka and India. Due to the sparseness of recent data in constructing a reasonable size corpus for computational modelling, our search time frame was expanded to approximately nine years from 2012-01-01 to 2021-10-31.

CrowdTangle supports keyword, hashtag, or URL search, combining with boolean search operators such as AND, OR, NOT. Our data collection process utilised the \textit{keywords} and \textit{phrases} option after a consultation with a clinical psychologist. Our keywords and phrases included "depression", "anxiety", "stress", "I feel unhappy", and 
“I feel like ending my life”.
To search Facebook posts in Sinhala language, these keywords and phrases were translated into Sinhala (See Appendix A.2 `Data Extraction' for the full list of keywords and phrases).  

We collected approximately 10,000 posts from Indian and Sri Lankan public Facebook groups. Each post includes metadata such as \textit{Group Id}, \textit{Group Name}, \textit{Text Post}, \textit{Post Created Time}, and Post Interaction (e.g., \textit{Like, Love}) Count. The extracted metadata did not contain any personal identification details such as Facebook user name or user Id. Therefore, Facebook user anonymity was preserved. During our thorough filtering process, we did not find any mention of Facebook user names inside post contents other than sentences like "\textit{please admin, approve this post etc.}". We recognised inherent demographic biases of data when the data extraction methodology disregards Facebook users' demographic information such as gender and age. We noticed a large amount of noise within the extracted Facebook data, resulting in difficulty in constructing a sufficiently large and demographically unbiased dataset.

\subsection{Data Cleaning}
Our data cleaning process included manually removing posts from inappropriate groups such as Facebook groups with adult content. We also excluded single sentence posts from the dataset using the NLTK tool\footnote{https://www.nltk.org/} since it is challenging to perform meaningful NLP processing to predict emotions from a single sentence. We also removed transliterated posts and translated all Sinhala language posts into English using Facebook Language Translator\footnote{https://developers.Facebook.com/docs/graph-api/reference/v12.0/app/translations}. Finally, we removed all the duplicate posts from the dataset. The data cleaning process resulted in a corpus of 2045 and 757 posts from Indian and Sri Lankan Facebook groups respectively (see Table\ref{tab:descriptive_stat} for `descriptive statistics' of the dataset).

\begin{table}
\centering
\begin{tabular}{lllll}
\hline
\textbf{Dataset} & \textbf{Sri Lankan} & \textbf{Indian} & \textbf{Full}\\
\hline
\small{\verb|Posts|} & \small{757} & \small{2045} & \small{2802} \\
\small{\verb|Sentences (ST)|} & \small{5827} & \small{9018} & \small{14845} \\
\small{\verb|ST per post|} & \small{12.1} & \small{4.9} & \small{--} \\
\small{\verb|Words per post|} & \small{188} & \small{93} & \small{--} \\
\hline
\end{tabular}
\caption{Descriptive Statistics of the filtered Facebook dataset}
\label{tab:descriptive_stat}
\end{table}


\subsection{Taxonomy Development}
As discussed in section \ref{sec:litrature}, the majority of research studies on emotion modelling rely on two popular taxonomies - Ekman's model \cite{ekman1992argument} and Plutchik's `Wheel of Emotions' \cite{plutchik1980general}.
Researchers tend to adapt these models 
by adding new emotions \cite{demszky2020goemotions} or removing emotions. Therefore, we adapted three basic emotions from these two models - \textit{fear}, \textit{anger}, and \textit{sadness} since empirical studies demonstrate that these three emotions are strongly associated with mental health issues. 
We also removed emotions such as \textit{disgust} and \textit{surprise} since they occurred infrequently in our selected data source. 

Our taxonomy development process adopted `open coding', a popular method in grounded theory to identify, describe or categorise phenomena found in qualitative data \cite{corbin1990grounded}. 
Firstly, we manually classified a random sample of 50 Facebook posts into meaningful categories (known as \textit{codes} \cite{miles1994qualitative}). To start with, we used the 3 basic emotions - \textit{fear}, \textit{anger}, and \textit{sadness} 
This analysis found additional mental states (e.g., \textit{suicidal thoughts}, \textit{loneliness}, and \textit{addictions}
that are likely associated with mental health conditions. Secondly, we consulted a clinical psychologist to refine the codes until we reached agreement on a taxonomy that contained \textit{codes} to annotate our dataset. After this consultation, we expanded the emotion of `fear' with `anxiety/stress' as these terms are used interchangeably in the Sri Lankan context. 
We also merged some codes 
due to their infrequent occurrence in the dataset (e.g., \textit{loneliness}) which could result in data sparseness when modelling. Accordingly, three additional codes were introduced as listed below (see Appendix A.1 for complete definitions of taxonomy);
\begin{itemize}
\item \textbf{Mental illness}: Posts that mention a diagnosis or a treatment related to a mental illness.
\item \textbf{Psychosomatic}: Posts on psychosomatic issues (e.g., fatigue, headaches) associated with an underlying mental condition.
\item \textbf{Other}: Posts that express a maladaptive mental condition but do not belong to any of the previously defined categories (e.g., addictions, loneliness).
\end{itemize}

Finally, we introduced '\textbf{irrelevant}' category if none of the above-defined codes were usable to annotate a particular post. Table \ref{tab:taxonomy_example} shows the finalised set of categories, along with examples, from our dataset. 

\begin{table*}
\centering
\begin{tabular}{p{2.6cm}p{13cm}}
\hline
\textbf{Category} & \textbf{Example} \\
\hline
\scriptsize{\verb|Mental illness (MI)|} & \scriptsize{I have been taking antidepressants since a long time, watching motivational videos, listening to relaxing music, but when things happen like a problem, my head is like a stone, why is that???}\\
\scriptsize{\verb|Sadness (SD)|} & \scriptsize{She has a lot of sadness in her heart because of a past incident for a long time.. she says it's hard to forget no matter how she tries.. she says she cannot live without forgetting that incident.. She says she is living because she cannot die}  \\
\scriptsize{\verb|Anxiety/Stress (AS)|} & \scriptsize{I'm so mentally down I'm in a lot of problems. I'm a person who has suffered a lot since I was a kid. I've never been loved even because of my family problems. From mom to dad because they separated when they were young. I lost everything I loved. I still suffer from that.}  \\
\scriptsize{\verb|Suicidal (SC)|} & \scriptsize{I'm suffering from depression \frownie Right now there are so many problems that are going on in my life. Sometimes I just want to end my life}  \\
\scriptsize{\verb|Anger (AG)|} & \scriptsize{Sometimes I think that I need to take revenge. Because revenge has been my addiction. If I don't take revenge, I'm in depression and so angry...I'm so afraid of myself because when I get angry I won't control and don't know what I have done.} \\
\scriptsize{\verb|Psychosomatic (PY)|} & \scriptsize{How to get good night sleep at night in depression? Suffering from Insomnia from last 3 months} \\
\scriptsize{\verb|Other (OT)|} & \scriptsize{I am studying and I feel lonely. Before some time when I worked I felt so excited and interested. But now no any interest and excitement} \\
\scriptsize{\verb|Irrelevant (IV)|} & \scriptsize{Anyone can love you when the sun is shining, but in the storms is where you'll learn who truly cares about you...} \\
\hline
\end{tabular}
\vspace{1ex}
{\raggedright \scriptsize{*Note - Due to sensitivity of data, we report an excerpt of the post} \par}
\caption{EmoMent Taxonomy and sample examples}
\label{tab:taxonomy_example}
\end{table*}

\subsection{Data Annotation}
We used three annotators to code the dataset. All of them were native Sri Lankans with masters-level experience in clinical psychology. They were recruited by distributing flyers within Psychology departments of three main universities and two higher educational institutes in Sri Lanka. They were employed as research assistants for 1.5 months and their time and effort were compensated based on the standard daily salaries in Sri Lanka. 
They were proficient in both Sinhala and English languages, and was familiar with Facebook mental health groups 
and cyber language. 
These annotators also had a sound understanding of South Asian culture and context.
None of the annotators is an author of this paper.

\begin{figure}
    \includegraphics[scale = 0.27]{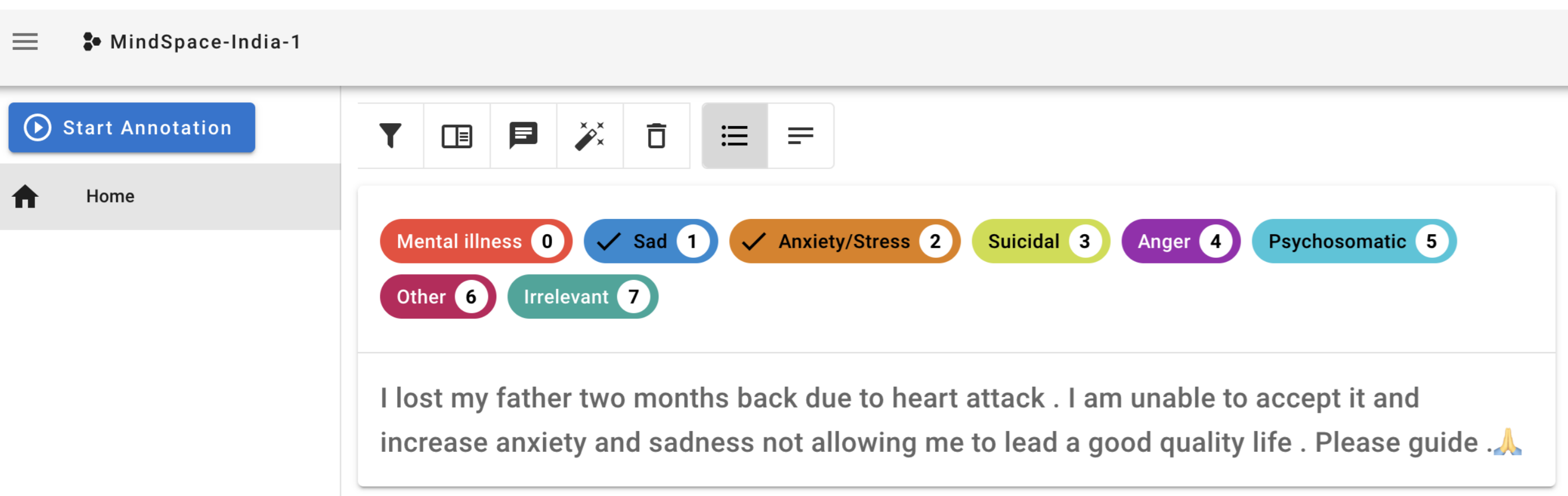}
    \caption{A screenshot of Doccano interface configured for annotators}
    \label{fig:doccano_interface}
\end{figure}

The task of an annotator was to read the entire post provided through Doccano web interface\footnote{https://github.com/doccano/doccano} and assign codes based on the taxonomy (Table \ref{tab:taxonomy_example}). Doccano is a popular web based, open source, text annotation tool. 
Figure \ref{fig:doccano_interface} shows an example of Doccano interface we configured for annotators. 
Each post could have one or more codes. However, the 'irrelevant' code was not allowed to be used jointly with other codes. 
Our annotation instructions emphasised the importance of making evaluations based on the information explicitly found in a given post without making assumptions. 
(see Appendix A.1 for more information about the `annotation guideline').

\section{Corpus Analysis}
\label{sec:analysis}
We constructed the EmoMent corpus by selecting posts which had two or more annotators agreeing on a category.

\subsection{Corpus Statistics}
Table \ref{corpus_stats} demonstrates corpus statistics of annotated EmoMent corpus. 
According to Table \ref{corpus_stats}, 
the majority of posts (62\%) had only one label, followed by 31\% of posts with two labels. Since 38\% of posts had more than one label, we have modelled this problem as a multi-label classification task (see section \ref{sec:modelling}). There were only 31 posts (i.e. 1\% of total posts) that had four or more labels.  According to the annotations, the most number of labels a post had were five and our dataset consists of eight such instances. The excerpt below demonstrates the five labels: \hlc[cyan]{\textit{mental illness}}, \hlc[pink]{\textit{anxiety/stress}}, \hlc[orange]{\textit{sadness}}, \hlc[lime]{\textit{suicidal}}, and \hlc[yellow]{\textit{anger}}.

\vspace{3mm}

\begin{small}
\noindent \textit{"[..] I'm posting this to find a solution because it's hard for me to bear. \hlc[cyan]{I'm in a depressed state. I took medication}. \hlc[pink]{I'm so nervous}. \hlc[orange]{Feeling sad}. I feel like dying. I just want someone to talk to me in the right words with love. \hlc[yellow]{Then my anger is going to calm down a little. I don't get angry for nothing. but for what she does. she lies to me}. I feel like her life was ruined because of me. It's too much pain to express when I feel like that. I feel like stabbing. Feeling so helpless. but it's hard for me to stay. Is there anyone who listens to me. Please help me. It's hard for me to live in this pain. Am I doing something wrong. I feel like I can't move forward. \hlc[lime]{I feel like there's no life} [..]"}

\end{small}


\begin{scriptsize}
\renewcommand{\arraystretch}{0.6}
\begin{table}
\centering
\begin{tabular}{ll}
\hline
\scriptsize{\verb|Number of (#) Posts|} & \scriptsize{2802} \\
\hline
\scriptsize{\verb|# Categories|} & \scriptsize{8}\\
\hline
\scriptsize{\verb|# labels per post|} & \scriptsize{1: 62\%, 2: 31}\%\\
 & \scriptsize{3: 6\%, 4 or more: 1}\%\\
\hline
\scriptsize{\verb|# posts where >2 annotators|} & \scriptsize{2106}\\
\scriptsize{\verb|agreed on at least 1 category|} & \\
\hline
\scriptsize{\verb|# posts where all 3 annotators|} & \scriptsize{1981}\\
\scriptsize{\verb|agreed on at least 1 category|} & \\
\hline
\scriptsize{\verb|# posts where annotators totally|} & \scriptsize{9}\\
\scriptsize{\verb|disagreed on at least 1 category|} & \\
\hline
\end{tabular}
\caption{EmoMent Corpus statistics}
\label{corpus_stats}
\end{table}
\end{scriptsize}

\begin{figure}
    \includegraphics[scale=0.39]{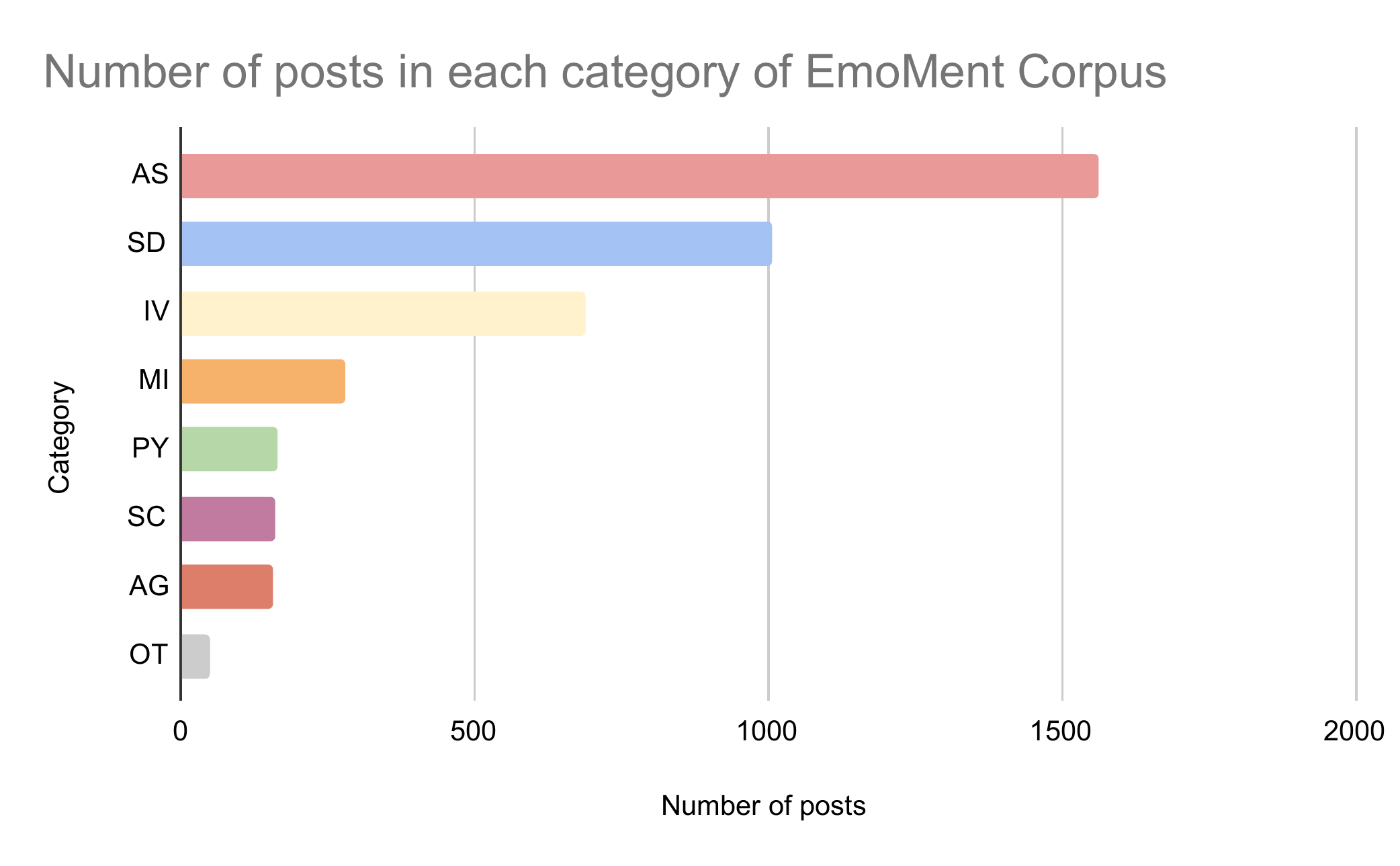}
    \caption{Number of posts in each label category, where at least two annotators agree for a particular label}
    \label{fig:bar_chart}
\end{figure}

Figure \ref{fig:bar_chart} shows the number of posts in each category, sorted  by the frequencies of the posts. 
According to Figure \ref{fig:bar_chart}, \textit{anxiety/stress} (AS) is the most common emotion (56\%) in the corpus, followed by \textit{sadness} (SD - 36\%). The majority of annotators (two or more) agreed that 24.5\% of posts in the corpus were \textit{irrelevant} (IV) based on our annotation guide.
Figure \ref{fig:bar_chart} shows a large disparity between the frequencies of AS (56\%) and PY (6\%), SC (6\%), AG (6\%). For example, \textit{anxiety/stress} was approximately nine times more frequent than \textit{suicidal thoughts}, demonstrating that social media users may express their anxiety/stress more frequently and openly than use social media as a platform to share their suicidal thoughts, 
This disparity in frequencies also led to a data imbalance problem when modelling. The \textit{other} (OT) category relates to 2\% of all annotated posts. We excluded OT from modelling since the purpose of this category was to identify potential other emotions that could be useful for future expansions of the corpus. However, we did not find any such emotions.

\subsection{Inter-Annotator Agreement}

In order to calculate the agreement between annotators, we used Fleiss' Kappa measurement \cite{fleiss1971measuring}. Fleiss' Kappa is used to determine the agreement when two or more annotators are present. 


\begin{table*}
\small
\centering
\begin{tabular}{llllllllll}
\hline
\textbf{Dataset} & \textbf{MI} & \textbf{SD} & \textbf{AS} & \textbf{SC} & \textbf{AG} & \textbf{PY} & \textbf{OT} & \textbf{IV} & \textbf{Average} \\
\hline
\verb|Sri Lankan| & 0.917 & 0.913 & 0.938 & 0.956 & 0.951 & 0.848 & 0.782 & 0.924 & 0.904 \\
\verb|Indian| & 0.794 & 0.831 & 0.714 & 0.834 & 0.810 & 0.660 & 0.516 & 0.783 & 0.743 \\
\verb|Average| & 0.856 & 0.872 & 0.826 & 0.895 & 0.880 & 0.754 & 0.649 & 0.853 & 0.823 \\
\hline
\end{tabular}
\caption{Inter-annotator agreement using Fleiss’ Kappa}
\label{tab:Kappa}
\end{table*}

According to Table \ref{tab:Kappa}, we have achieved a ‘very good’ inter-annotator agreement for the Sri Lankan dataset with Fleiss’s Kappa of 0.9, enabling a promising human agreement for computational modelling. Almost all the label categories except ‘other’ have obtained over 0.8 of agreement. Additionally, the Indian dataset also achieved a ‘good’ Kappa value of 0.74. It is expected that a higher inter-annotator agreement for the Sri Lankan dataset was obtained as compared to the Indian dataset since the annotators were native Sri Lankan domain experts who have a better contextual knowledge about mental health issues among Sri Lankans, than among Indians. Table \ref{tab:Kappa} shows that \textit{anger} and \textit{suicidal} have the highest and \textit{other} and \textit{psychosomatic} have the lowest agreement respectively. Interestingly, the highest annotator agreements were observed from most infrequent categories - \textit{anger} and \textit{suicidal} (see Figure \ref{fig:bar_chart}).



\section{Modelling}
\label{sec:modelling}

\subsection{Data pre-processing}
In order to prepare \textit{EmoMent} dataset for downstream modelling tasks, we first associate each post x\textsuperscript{i} with a binary vector $y\textsuperscript{i} = [y\textsubscript{1}\textsuperscript{i} , . . . , y\textsubscript{k}\textsuperscript{i} ] \in \{0, 1\}\textsuperscript{k}$, where k represents the number of distinct labels in the taxonomy.
Here y\textsubscript{j}\textsuperscript{i} is assigned 1 if and only if the post x\textsuperscript{i} is associated with the label j. We determine whether the post x\textsuperscript{i} is associated with the label j based on whether 2 or more annotators agree with the association. We removed posts which were not associated with any label to yield our final dataset, referred to as EmoMent\textsubscript{all}. Additionally, we created a secondary dataset referred to as EmoMent\textsubscript{relevant}, selecting posts which are not associated with the label `\textit{Irrelevant} (IV)’. Hence, EmoMent\textsubscript{relevant} is a subset of EmoMent\textsubscript{all}. We randomly split each dataset into training, validation and test splits in 70:15:15 ratio (see Appendix \ref{app:data_splits} for detailed dataset split).

\subsection{Emotion-annotated Mental Health Models}
We propose experimental baselines for two associated tasks. The `first task' is a binary classification task of determining whether a post is \textit{relevant} or \textit{irrelevant} to a mental health condition. The `second task' is a multi-label classification task of associating correct labels (e.g., MI, SD, AS, SC)  with a given post. 
As discussed in section \ref{sec:analysis}, we chose not to consider the OT category for modelling.

We use BERT \cite{devlin2018bert} and RoBERTa \cite{liu2019roberta} pre-trained language models. Our selection of BERT-based pre-trained models were motivated by previous impressive performance across different NLP tasks and related studies \cite{demszky2020goemotions} that used BERT-based models to propose strong baselines. RoBERTa is an optimised model based on BERT, and it has shown to outperform BERT in numerous tasks \cite{liu2019roberta}. Hence, we developed strong baseline models using both BERT and RoBERTa.

\begin{figure*}
    \centering
    \includegraphics[scale=0.4]{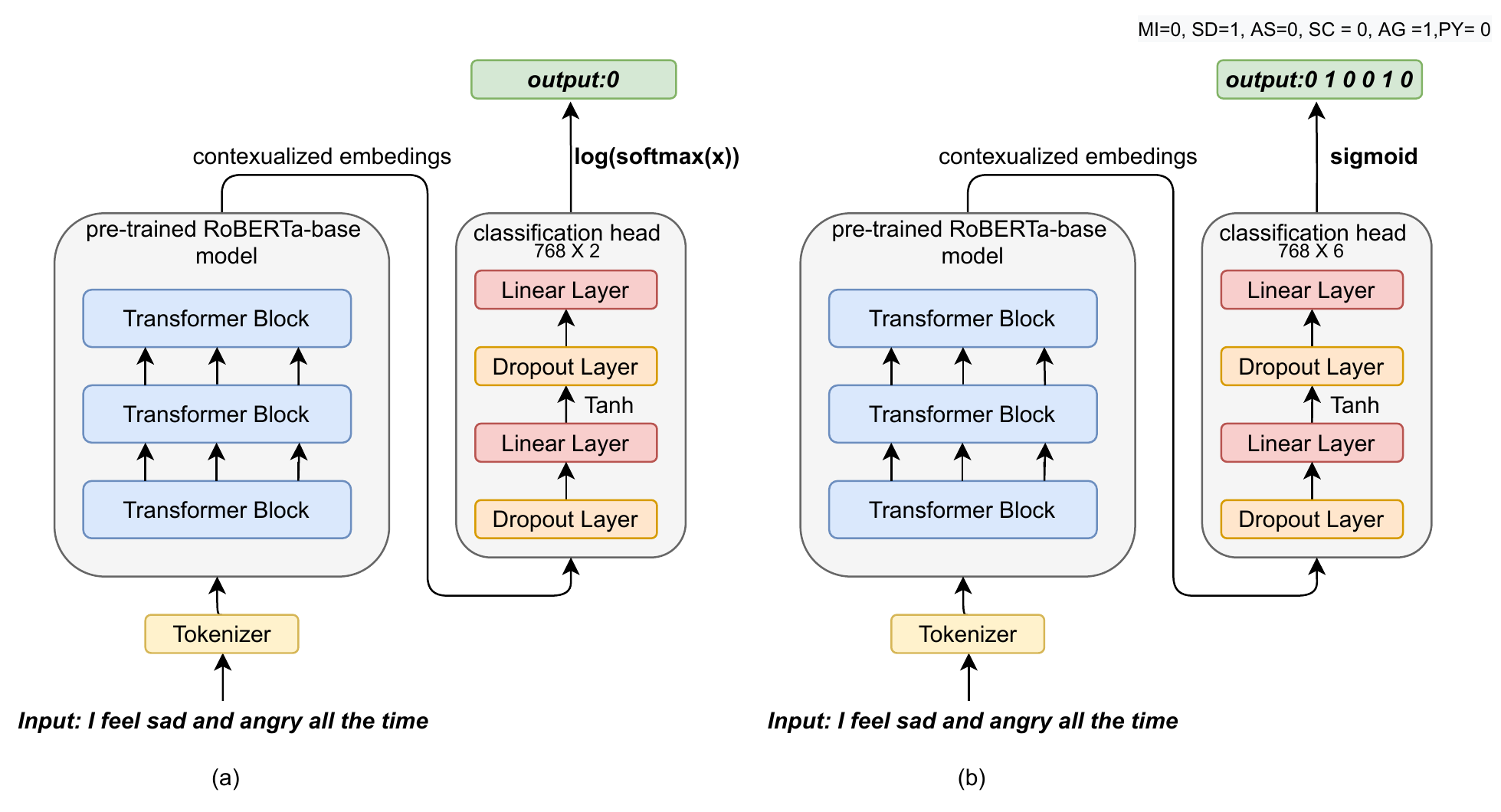}
    \caption{Architecture of the RoBERTa based models}
     \label{fig:arc_diagram}
\end{figure*}

Figures \ref{fig:arc_diagram}(a) and \ref{fig:arc_diagram}(b) show the architecture of our RoBERTa based 
binary and multi-label classification models respectively. 
We use the Pytorch HuggingFace library\footnote{https://huggingface.co/} to implement the models. We ran all our experiments on the default GPUs provided by Google Colab\footnote{https://colab.research.google.com/}.

\subsubsection{Binary Classification Task}
To address the `first task', we fine tune BERT and RoBERTa based models on the training split (70\%) of the EmoMent\textsubscript{all} dataset. Our hyper-parameters tuning and performance evaluation used validation (15\%) and test (15\%) splits of the EmoMent\textsubscript{all} dataset.
To selected the best hyper-parameter combination, we trained three different models per each hyper-parameter combination by only modifying the random seed values, and compared the average scores obtained.  We used cross-entropy loss as the loss function during training. 
 
First we fine tuned a bert-base-cased model with a classification head on top for the binary classification task. 
During our experiments, we change the hyper-parameters \textit{learning rate}, \textit{batch size} and the \textit{number of epochs}. We set the \textit{warmup ratio} to 0.1 and keep the default values provided in HuggingFace implementation for the rest of the hyperparameters. We observe the best results when we select a learning rate of 2e-05, a batch size of 8, and train the model for 5 epochs.

Next we fine tune a roberta-base model with a classification head on top for the binary classification task. To finetune the RoBERTa model, we followed a strategy similar to BERT. However, we observed best results when we set the learning rate to 2e-05, batch size to 8 and train the model for 3 epochs. We found that after 3 epochs, models tend to get overfitted to the training dataset.

\subsubsection{Multi-label Classification Task}
To address the `second task', we fine tune BERT and RoBERTa based models on the training split (70\%) of the EmoMent\textsubscript{relevant} dataset 
Our BERT model was a bert-base-cased pre-trained model with a classification head on top. Similarly, our RoBERTa model was a roberta-base pre-trained model with a classification head on top. The output size of the last linear layer of both of these models is set to 6 since we only considered the categories MI, SD, AS, SC, AG and PY for this task. 

We used a binary cross-entropy loss function during training. To mitigate the negative impact from class imbalance, we input a vector of positive class weights to the loss function to be used when computing the loss. We computed this weight vector using the training split of the EmoMent\textsubscript{relevant} dataset.  For each label, we divided the number of negative training data instances associated with it by the number of positive training data instances associated with it, and rounded it off to the nearest integer. If the number of negative training data instances was less than the number of positive training data instances, we assigned a default positive class weight of 1.

We used the validation split (15\%) and test split (15\%)  of the EmoMent\textsubscript{relevant} dataset) to tune hyper-parameters and evaluate the models respectively. We experimented by adjusting the \textit{learning rate}, \textit{batch-size} and the \textit{number of epochs}. While fine tuning BERT and RoBERTa models, we set the warmup ratio to 0.1 and kept the default values provided in the HuggingFace implementation for the rest of the hyperparameters. As similar to binary classification problem, for each hyper-parameter combination, we trained 3 separate models by updating the random seed values, and compare the average scores obtained. We find that both BERT and RoBERTa based models perform well when we use a learning rate of 2e-05, a batch-size of 8 and train the model for 5 epochs. 
We report the precision, recall and the F1 score of each label separately, without averaging the results across labels (Table \ref{tab:multi}).

\subsection{Results}

We have summarised the results in Tables \ref{tab:binary} \& \ref{tab:multi}. Since we trained 3 models for each hyper-parameter configuration by updating the random seed value, the results we have reported are the \textit{macro-averaged} scores.

We observed that the RoBERTa model performs better than the BERT model in both tasks. In the first task, the RoBERTa model achieved an average F1 score of 0.76 compared to the BERT model which achieved an average F1 score of 0.72. In the second task, the RoBERTa model achieved a macro-averaged F1 score of 0.77 compared to the macro-averaged F1 score of 0.71 achieved by BERT. 

For the multi-label classification task, we have also reported F1 scores of the individual categories. We have observed that both BERT and RoBERTa models report the lowest F1 score for the PY category. From the Table \ref{tab:Kappa}, we observed that the PY category has a relatively lower inter-annotator agreement compared to MI, SD, AS, SC and AG categories. It is likely that this higher variability of data associated with the PY category could have caused both BERT and RoBERTa models to perform poorly.

We extracted misclassified posts by the best performing models for further analysis (see Table \ref{tab:misclassified} of Appendix A.4 for a sample of misclassified posts). We observed that when classifying posts that seek general information or offer advice on mental health conditions, RoBERTa based binary classification model tends to get confused at times (see first 2 examples on \textit{relevant/irrelevant} in Table \ref{tab:misclassified}).  In the case of the multilabel-classification task, we observed that certain labels like PY gets misclassified more often.  As noted in Table \ref{tab:Kappa}, the inter-rater agreement for the PY category is relatively low, and it is likely that the lower agreement has contributed to the misclassification of the PY category.



\subsection{Limitations}

As described in section 3.2, we first translated the extracted posts from Sinhala language to English prior to annotating the data. Translating the posts to English makes the dataset accessible to a much broader research community. We acknowledge that translating posts in this manner can lead to biased results. This is a limitation of the current corpus. However we argue that the benefits of translating the posts to English outweigh the disadvantages.

In this study we limited our focus to two countries in the South Asian region, Sri Lanka and India. Thus, our corpus is not representative of all the demographics in the world, and we acknowledge this as a limitation. However, we believe this does not diminish the usefulness of the corpus. The South Asian region is a populous region with more than  20\% of the world’s population  \cite{veron2008demography}. Therefore,  we believe our work would be beneficial to a large audience.

\begin{table}
\centering
\begin{tabular}{llll}
\hline
\textbf{Model} & \textbf{Precision} & \textbf{Recall} & \textbf{F1-score}\\
\hline
\verb|BERT| & 0.79 & 0.67 & 0.72 \\
\verb|RoBERTa| & 0.84 & 0.71 & 0.76 \\
\hline
\end{tabular}
\caption{Results from the binary classification task}
\label{tab:binary}
\end{table}

\begin{small}
\begin{table}
\centering
\begin{tabular}{lllllll}
\hline
\textbf{\small{Label}}
&\multicolumn{2}{c}{\textbf{Precision}}
& \multicolumn{2}{c}{\textbf{Recall}}
& \multicolumn{2}{c}{\textbf{F1 Score}}\\
& \textbf{\tiny{BERT}} & \textbf{\tiny{RoBERTa}} 
& \textbf{\tiny{BERT}} & \textbf{\tiny{RoBERTa}}
& \textbf{\tiny{BERT}} & \textbf{\tiny{RoBERTa}}\\
\hline
\verb|MI|& 0.7 & 0.77 &	0.66 &	0.76 &	0.68 &	0.76\\
\verb|SD|& 0.84	& 0.85 & 0.84 &	0.88 &	0.84 &	0.87 \\
\verb|AS|& 0.81 & 0.85 & 0.94 &	0.94 &	0.87 &	0.89 \\
\verb|SC|& 0.66	& 0.76 &	0.78 &	0.77 &	0.72 &	0.76 \\
\verb|AG|& 0.6 &	0.66 &	0.85 &	0.83 &	0.7 &	0.73 \\
\verb|PY|& 0.42 &	0.5 &	0.54 &	0.71 &	0.47 &	0.59 \\
\hline
\textbf{\scriptsize{macro}} & \textbf{0.67} & \textbf{0.73} & \textbf{0.77} & \textbf{0.82} & \textbf{0.71} & \textbf{0.77} \\
\hline
\end{tabular}
\caption{Results from the multi-label classification task}
\label{tab:multi}
\end{table}
\end{small}

\section{Conclusion}
This paper presented the first emotion-annotated mental health corpus - \textit{EmoMent}, which was developed using Facebook posts from two South Asian countries - Sri Lanka and India. We have provided a comprehensive research study, demonstrating the development of an empirically-sound emotion-annotated mental health taxonomy using the grounded theory approach. 

We also developed strong baselines using RoBERTa-based models and achieved an F1 score of 0.76 for the \textit{first task} (i.e., predicting the relevance of a post to a mental health condition) and a macro-averaged F1 score of 0.77 for the \textit{second task} (i.e., predicting the relevant labels in our taxonomy). However, our results suggest that there is ample room for future improvements in emotion-annotated mental health modelling. The models presented in the paper consider the emotion-annotated mental health modelling as two separate tasks, one binary classification to determine the relevancy, and multi-label classification to predict fine-grained labels of posts. An interesting next step would be to co-model these two tasks by leveraging multi-task learning (MTL).

\section{Ethical Considerations}

We curated EmoMent corpus from publicly available Facebook posts while adhering to the data policy of Meta Platforms Inc.\cite{MetaUserPolicy}, the parent organization of Facebook and CrowdTangle.

 During data collection we took steps to filter out personally identifiable information (see section 3.1). However, we acknowledge the possibility of tracing back the origins of these posts since the original posts are available in the public domain.  We further acknowledge that provided annotations increase the sensitivity of the dataset.  Therefore, to reduce the risk of data misuse, when releasing the dataset for academic research upon request, we plan to do so under a strict confidentiality agreement.

We also acknowledge that all mental health related diagnoses must be made only by qualified mental health practitioners, and that the computational models proposed in this study cannot be used to make such diagnostic claims about a patient.

\section*{Acknowledgements}
Authors would like to acknowledge the Australian Academy of Science and the Australian Department of Industry, Science, Energy and Resources for providing financial support to conduct this research under their ‘Regional Collaborations Programme
COVID-19 Digital Grants’. We would also like to thank the three annotators for their support and we are grateful to the CrowdTangle research team for providing us a platform to collect Facebook data.


\bibliography{anthology,custom}
\bibliographystyle{acl_natbib}

\appendix

\section{Appendix}
\label{sec:appendix}

\subsection{Annotation Guideline: Taxonomy}

Table \ref{tab:appendix_taxonomy} provides definitions per each category included in the annotation guideline.

\begin{table*}
\centering
\begin{tabular}{p{2.6cm}p{13cm}}
\hline
\textbf{Category} & \textbf{Definition} \\
\hline
\scriptsize{\verb|Mental illness (MI)|} & \scriptsize{Posts that explicitly mention a diagnosis of a mental illness or getting treatments for a mental illness such as depression, anxiety and seek help. Posts that expresses self-identification of mental illness may be due to history of treatments.}\\
\scriptsize{\verb|Sadness (SD)|} & \scriptsize{Posts that express sadness, unhappy or sorrow that may lead to a maladaptive mental condition or mental illness.}  \\
\scriptsize{\verb|Anxiety/Stress (AS)|} & \scriptsize{Posts that express stress, fear or worry about something (e.g. past, future, physical appearance, religious beliefs) using the words such as anxiety, worry, fear, stress that may lead to a maladaptive mental condition or mental illness.}  \\
\scriptsize{\verb|Suicidal (SC)|} & \scriptsize{Posts that express suicidal thoughts, no interest in life (e.g. I feel like taking my own life).}  \\
\scriptsize{\verb|Anger (AG)|} & \scriptsize{Posts that express anger using words such as anger that may lead to a maladaptive mental condition or mental illness.} \\
\scriptsize{\verb|Psychosomatic (PY)|} & \scriptsize{Posts that express psychosomatic issues (e.g. insomnia, fatigue, headaches, upset stomach) that associated with underlying mental distress or may lead to a maladaptive mental condition or mental illness.} \\
\scriptsize{\verb|Other (OT)|} & \scriptsize{Posts that may lead to a maladaptive mental condition or mental illness but do not belong to any of the above categories (e.g., addictions, loneliness, social skill deficits such ascommunication issues, problem solving issues, interpersonal issues).} \\
\scriptsize{\verb|Irrelevant (IV)|} & \scriptsize{Posts that seek information on matters related to mental conditions but do not discuss about an issue of the poster or a third party. Posts that thank others who helped. Matters related to social media group (e.g. rules of the Facebook group, objectives). Posts written using languages other than Sinhala or English. } \\
\hline
\end{tabular}
\vspace{1ex}
\caption{EmoMent Taxonomy and definitions}
\label{tab:appendix_taxonomy}
\end{table*}

\subsection{Data Extraction}
\begin{figure}
    \centering
    \includegraphics[keepaspectratio=true, width=0.575\textwidth]{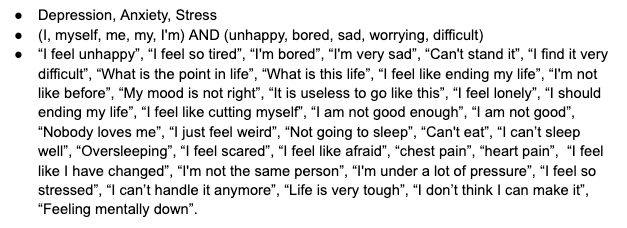}
    \caption{English search keywords and phrases used in the data extraction}
    \label{fig:english_search_phrases}
\end{figure}

\begin{figure}
    \centering
    \includegraphics[keepaspectratio=true, width=0.575\textwidth]{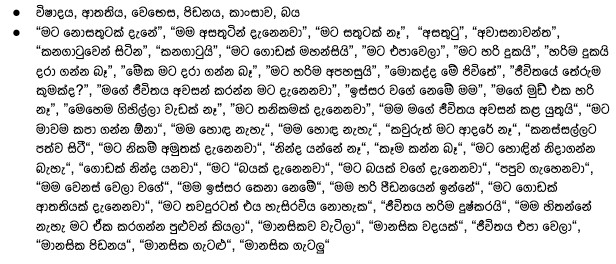}
    \caption{Sinhala search keywords and phrases used in the data extraction}
    \label{fig:sinhala_search_phrases}
\end{figure}

Figures \ref{fig:english_search_phrases} and \ref{fig:sinhala_search_phrases} show the English and Sinhala search keywords and phrases used in the data extraction.

\subsection{Composition of Training, Evaluation and Test Datasets}
\label{app:data_splits}

Table \ref{tab:dataset_splits} demonstrates the percentages of positive and negative instances associated with each label in training, validation and test splits.
\begin{table*}[h]
\centering
\begin{tabular}{p{2cm}p{2cm}p{2cm}p{2cm}p{2cm}p{2cm}p{2cm}}

\textbf{Label}
& \multicolumn{2}{c}{\textbf{Train}}
& \multicolumn{2}{c}{\textbf{Validation}}
& \multicolumn{2}{c}{\textbf{Test}}\\
&  \textbf{1} & \textbf{0} & \textbf{1} & \textbf{0} & \textbf{1}  & \textbf{0} \\
\hline
\verb|MI|& 14\% & 86\% & 11\% & 89\% & 15\% & 85\%\\
\verb|SD|& 47\% & 53\% & 51\% & 49\% & 49\% & 51\% \\
\verb|AS|& 74\% & 26\% & 75\% & 25\% & 73\% & 27\% \\
\verb|SC|& 8\% & 92\% & 6\% & 94\% & 8\% & 92\% \\
\verb|AG|& 8\% & 92\% & 6\% & 94\% & 7\% & 93\% \\
\verb|PY|& 7\% & 93\% & 9\% & 91\% & 8\% & 92\% \\
\hline
\end{tabular}
\caption{Percentages of positive and negative instances associated with each label in training, validation and test splits of EmoMent\textsubscript{relevant} dataset.
}
\label{tab:dataset_splits}
\end{table*}

\subsection{A Sample of Misclassified Posts}

Table \ref{tab:misclassified} shows a sample of posts misclassified by the models

\begin{table*}
\centering
\begin{tabular}{p{10.2cm}p{2cm}p{1cm}p{1cm}}
\hline
\textbf{Post} & \textbf{Predicted} & \textbf{Actual} & \textbf{Misclassified} \\
\hline 
{\scriptsize{I am taking meditation classes for stress anxiety and depression... Timing is morning if interested so reply}} & \scriptsize{IV=0} & \scriptsize{IV=1} & \scriptsize{IV} \\
\hline 
\scriptsize{Can someone tell me the best meditation for anxiety relief..It will be of great help} 
 & \scriptsize{IV=0} & \scriptsize{IV=1} & \scriptsize{IV} \\
\hline
\scriptsize{Anxiety is off the charts Everytime I doze off I am woken up my a feeling that I am falling and I can't breathe hate this feeling do now???} 
 & \scriptsize{AS=1, PY=1} & \scriptsize{AS=1} & \scriptsize{PY} \\
\hline
\scriptsize{Just woke up with a bad  dream.people are killing each other,there are hail storms, something is coming from sky destroying the earth,my family is pushing me for marriage.Since then my heart is racing}
& \scriptsize{AS=1} & \scriptsize{PY=1} & \scriptsize{PY} \\
\hline
\end{tabular}
\caption{A sample of misclassified posts}
\label{tab:misclassified}
\end{table*}

\end{document}